\documentclass[letterpaper, 10pt, conference, twoside]{style/ieeeconf}    

\makeatletter
\let\NAT@parse\undefined
\makeatother

\usepackage{dblfloatfix}

\usepackage[numbers,sectionbib,sort&compress]{natbib}
\usepackage{bm}
\usepackage{gensymb}
\usepackage{xcolor}
\usepackage{graphicx}
\usepackage{amsmath}
\usepackage{amssymb}
\usepackage{subcaption}
\usepackage{amsfonts}
\usepackage{siunitx}
\usepackage{booktabs}
\usepackage{makecell}
\usepackage{multirow}
\usepackage{upgreek}
\usepackage[font=small]{caption}
\usepackage[export]{adjustbox}
\usepackage{tikz}
\usepackage{tabularx}
\usepackage{svg}
\usepackage{float}
\usepackage{sidecap} \sidecaptionvpos{figure}{c}
\usepackage{hyperref}
\hypersetup{colorlinks,breaklinks,
linkcolor=[rgb]{0.5,0.,0.},
citecolor=[rgb]{0.000,0.427,0.173},
urlcolor=[rgb]{0.031,0.318,0.612}}

\usepackage[nameinlink, capitalize]{cleveref}
\usepackage{color}
\definecolor{CommentPink}{rgb}{1,0.2,0.5}
\definecolor{CommentBlue}{rgb}{0,0,1}
\definecolor{CommentGreen}{rgb}{0,1,0}

\Crefname{section}{Sec.}{Sec.}
\Crefname{equation}{Eq.}{Eq.}

\newcommand\blfootnote[1]{%
\begingroup 
\renewcommand\thefootnote{}\footnote{#1}%
\addtocounter{footnote}{-1}%
\endgroup 
}

\newcommand{\etal}{\textit{et al}.}
\newcommand{\ie}{\textit{i}.\textit{e}., }
\newcommand{\eg}{\textit{e}.\textit{g}., }

\DeclareMathOperator*{\argmin}{argmin}

\usepackage[printonlyused,withpage,nolist,nohyperlinks]{acronym}

\IEEEoverridecommandlockouts   

\title{\LARGE \bf How Many Views Are Needed to Reconstruct \\ an Unknown Object Using NeRF?}

\author{Sicong Pan$^{\star}$ \and Liren Jin$^\star$ \and Hao Hu \and Marija Popović \and Maren Bennewitz %
}

\begin{document}

\twocolumn[{%
\renewcommand\twocolumn[1][]{#1}%
\maketitle

\vspace{-0.5cm}

\begin{figure}[H]
\hsize=\textwidth
\centering
\includegraphics[width=0.8\textwidth]{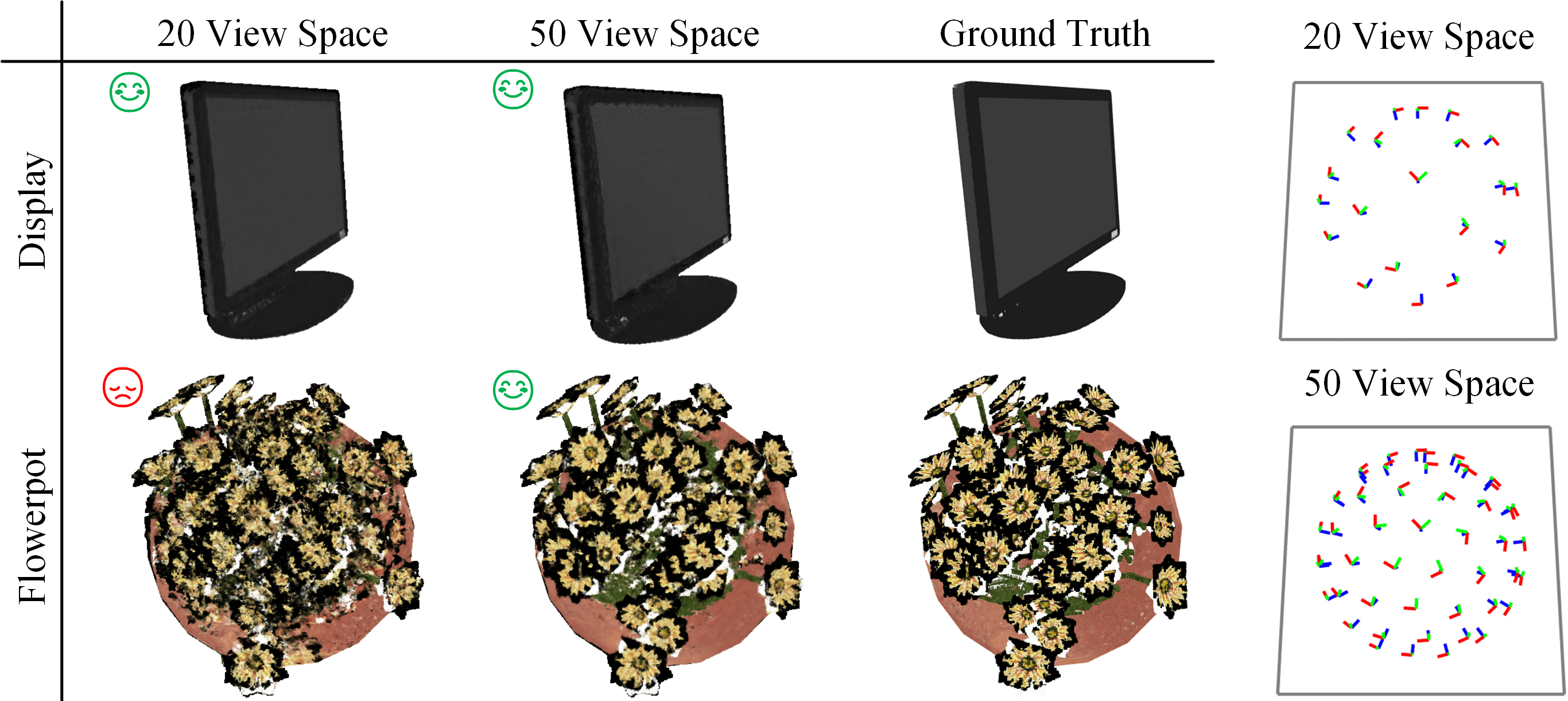}
\caption{
An example of how object complexity affects the required number of views to reconstruct an unknown object using NeRF. The objects are trained under 20 and 50 views of hemispherical view spaces shown in the last column. The images rendered from novel test views are shown in the first two columns. As can be seen, a less colorful and geometrically simple display can be reconstructed well with 20 views, whereas a colorful and geometrically complex flowerpot requires 50 views to achieve a good result. In this work, we present an approach to predict the required number of views by a deep neural network based on the complexity of the object to be reconstructed.
} 
\label{fig_cover}
\end{figure}
}]

\thispagestyle{empty} 
\pagestyle{empty}

\begin{abstract} 
Neural Radiance Fields (NeRFs) are gaining significant interest for online active object reconstruction due to their exceptional memory efficiency and requirement for only posed RGB inputs. Previous NeRF-based view planning methods exhibit computational inefficiency since they rely on an iterative paradigm, consisting of (1)~retraining the NeRF when new images arrive; and (2)~planning a path to the next best view only. To address these limitations, we propose a non-iterative pipeline based on the Prediction of the Required number of Views (PRV). The key idea behind our approach is that the required number of views to reconstruct an object depends on its complexity. Therefore, we design a deep neural network, named PRVNet, to predict the required number of views, allowing us to tailor the data acquisition based on the object complexity and plan a globally shortest path. To train our PRVNet, we generate supervision labels using the ShapeNet dataset. Simulated experiments show that our PRV-based view planning method outperforms baselines, achieving good reconstruction quality while significantly reducing movement cost and planning time. We further justify the generalization ability of our approach in a real-world experiment.
\end{abstract} 

\blfootnote{$^\star$These authors contributed equally to this work.}%
\blfootnote{Sicong Pan and Maren Bennewitz are with the Humanoid Robots Lab, Liren Jin and Marija Popović are with the Institute of Geodesy and Geoinformation, University of Bonn, Germany. Maren Bennewitz is additionally with the Lamarr Institute for Machine Learning and Artificial Intelligence, Germany. Hao Hu is with Intel Asia-Pacific Research \& Development Ltd. This work has partially been funded by the Deutsche Forschungsgemeinschaft (DFG, German Research Foundation) under grant 459376902 – AID4Crops and under Germany’s Excellence Strategy, EXC-2070 – 390732324 – PhenoRob. Corresponding author: \href{mailto:span@uni-bonn.de}{span@uni-bonn.de}
}%


\section{Introduction}
\label{sec_intro}

Object 3D reconstruction is a crucial task in robotic active vision~\cite{chen2011active}, which utilizes online view planning to move the camera to maximize the information about the object to be reconstructed. Prior research mostly uses explicit 3D representations such as point clouds~\cite{zeng2020pc,border2022surface}, voxel grids~\cite{delmerico2018comparison,pan2022aglobal}, and meshes~\cite{wu2014quality,lee2020automatic} to perform view planning. However, these methods require discretizing the scene thus leading to substantial memory consumption. Meanwhile, updating an explicit representation relies on depth-sensing modalities, \ie fusion by depth. In contrast, Neural Radiance Fields (NeRFs)~\cite{mildenhall2021nerf} offer an alternative approach by implicitly modeling 3D space from a set of posed RGB images, utilizing continuous functions implemented as deep neural networks, \ie fusion by learning. Consequently, NeRFs demonstrate memory efficiency and provide high reconstruction quality. With the emergence of highly efficient training architectures like Instant-NGP~\cite{muller2022instant}, the integration of NeRF models into online view planning becomes feasible.

Existing NeRF-based view planning methods follow the active learning paradigm~\cite{pan2022activenerf} to achieve good reconstruction performance, in which the most informative next-best-view (NBV), \eg the most uncertain view, is selected iteratively. The robot navigates to the NBV for new image collection until a predefined maximum number of iterations, \ie the required number of views, or a performance plateau is reached. The primary limitation lies in the fact that these methods rely on greedy NBV planning based on the current NeRF state. This often necessitates retraining the NeRF when new images are collected, leading to computational inefficiency compared to faster depth fusion updates. Another drawback is that the robot only iteratively executes paths between NBVs resulting in path planning inefficiency. Motivated by the inherent inefficiencies, we  develop a novel online view planning method that discards the need for iterative planning


To realize this capability, two essential components are required to plan all views at once: (1)~the required number of views until the reconstruction mission can be terminated; (2)~the view configuration, \ie how and where to place these views. Regarding the view configuration, we assume a hemispherical view space and simply utilize the solution to the Tammes problem~\cite{lai2023iterated}, which finds the placement of a given number of points on a sphere to maximize the minimum distance between them. Although the theoretically optimal views should be adaptively configured based on the specific object to be reconstructed, our experimental results suggest that using the Tammes configuration is sufficiently effective. In this work, our primary focus is on discussing the problem of finding the required number of views to reconstruct a specific object.

This problem is not fully discussed in previous NeRF-based view planning literature, which relies on purely heuristic approaches or a user-defined number of views~\cite{lin2022active,jin2023neu,ran2023neurar,sunderhauf2023density}. These methods cannot guarantee both an adequate result and a highly efficient reconstruction. In particular, complex objects usually require denser views to achieve high reconstruction quality, while less views are sufficient to reconstruct simple objects. As shown in Fig.~\ref{fig_cover}, different objects have different levels of complexity, such as color, geometry, etc., and require different numbers of views to achieve a good reconstruction. Based on this observation, our novel method proposes to predict the object-specific required number of views to strike a balance between quality and efficiency in active NeRF reconstruction.

We model the relationship between the object complexity and the required number of views as a regression problem solved by a deep neural network PRVNet. We devise our PRVNet to extract features from multiple RGB images captured from initial views, thereby fostering a comprehension of the object complexity. To supervise PRVNet training, we generate a new dataset with different objects from ShapeNet~\cite{shapenet2015} labeled with the required number of views. The label of the required number of views is computed by finding the minimum number of views to reach a prefixed gradient threshold of the curve representing Peak Signal-to-Noise Ratio (PSNR) performance over the number of views of a specific object. Given the number of views predicted by PRVNet, our method configures a Tammes view space and computes globally shortest paths between these views. This enables us to reduce the movement cost in contrast to iterative methods that only plan a path to the NBV.

Compared to two baselines from recent literature~\cite{lin2022active,sunderhauf2023density}, our PRV-based view planning can reconstruct an unknown object with better or comparable NeRF representation with significantly less movement cost and planning time. The contributions of our work are threefold:
\begin{itemize}
\item An efficient pipeline for active NeRF reconstruction, avoiding iterative planning with time-consuming retraining and high movement cost.
\item An unknown object reconstruction method based on the prediction of the required number of views, which balances between the quality and efficiency of reconstruction.
\item Our PRVNet along with a dataset containing the required number of views for every object, modeling the relationship between object complexity and required number of views in NeRFs.
\end{itemize}
To support reproducibility, our implementation and dataset is published at \url{https://github.com/psc0628/NeRF-PRV}.

\section{Related Work}\label{S: related work}

\subsection{View Planning for Object Reconstruction} \label{SS:view planning for object reconstruction}

View planning methods for object reconstruction can be largely grouped into two classes: search and learning. Zeng \etal~\cite{zeng2020view} summarized modern search-based methods as a generate-and-test procedure that generates a set of candidate views and tests each view by its current utility. The utility is commonly defined by intuitive concepts such as frontiers~\cite{border2018surface,border2020proactive,zaenker2021combining}, shape analysis~\cite{wu2014quality,lee2020automatic,menon2022viewpoint}, occupancy with occlusion awareness~\cite{daudelin2017adaptable,delmerico2018comparison,zaenker2021viewpoint}, and global coverage optimization~\cite{pan2022aglobal,pan2023global,zaenker2023graph}. Deep learning-based methods treat the NBV planning problem as a classification or regression problem, which trains a network given candidate view and its potential surface coverage value as the label~\cite{mendoza2020supervised,zeng2020pc,vasquez2021next,han2022double}. Some classification networks~\cite{pan2022scvp,pan2023one} output multiple views at once by learning from set-covering optimization problems. Other methods formulate a reinforcement learning framework~\cite{peralta2020next,zeng2022deep,dengler2023viewpoint} to learn a view planning policy from rewards in the environment.

The stopping criterion, \ie required number of views, has recently attracted the interest of researchers as it determines the stopping time and the efficiency of the reconstruction. Delmerico \etal~\cite{delmerico2018comparison} propose stopping the reconstruction when any candidate view falls below a user-defined threshold. Yervilla-Herrera \etal~\cite{yervilla2019optimal,yervilla2022bayesian} terminate the reconstruction when the number of frontier voxels is not changing (or equivalently the entropy of the frontier voxels is constant or the variation is smaller than a threshold). Pan \etal~\cite{pan2022scvp} utilize a deep neural network to output a set of views and use the size of the predicted view set as the required number of views. However, these studies consider an explicit map representation. Defining the stopping criterion for implicit representations remains an open problem.

\subsection{Next-Best-View Planning in Neural Radiance Fields} \label{SS:next-best view planning in neural radiance fields}

Different from previous works using explicit map representations, NeRF-based methods pose challenges in quantifying the utility of the view candidates. Since the explicit geometry is not directly available from NeRFs, view selection based on surface coverage or frontiers is hard to achieve. Emerging works study NeRF-based view planning by incorporating uncertainty quantification into NeRF representations. Pan \etal~\cite{pan2022activenerf} learn NeRFs assuming Gaussian distribution on the radiance value and train the variance prediction by minimizing negative log-likelihood. This work adds the view candidate with the highest information gain, \ie the highest uncertainty reduction, to the existing training data. Instead of learning uncertainty prediction additionally, Lee \etal~\cite{lee2022uncertainty} exploit the entropy of the density prediction along the ray as an uncertainty measure with respect to the scene geometry. The entropy is used to guide measurement acquisition towards geometrically less precise or unexplored parts. Thanks to the recent development of fast training by Instant-NGP \citep{muller2022instant}, Lin \etal~\cite {lin2022active} and S{\"u}nderhauf \etal~\cite{sunderhauf2023density} train an ensemble of NeRF models for a single scene and treat variance of the ensemble’s prediction as uncertainty quantification. Jin \etal~\cite{jin2023neu} train a generalizable image-based neural rendering network together with uncertainty prediction with respect to the input data uncertainty. 

The above-mentioned works focus on uncertainty quantification in NeRFs and use it for NBV planning. A key assumption is that the required number of views is defined by a user, \eg common choices are 10 and 20 total views~\citep{lin2022active,zhan2022activermap,jin2023neu}. Ran \etal~\cite{ran2023neurar} stop at 28 views and also tests 18, 38, and 58 views. Lee \etal~\cite{lee2022uncertainty} use 15 clustered views in the real world for initialization and then plan 12 NBVs. S{\"u}nderhauf \etal~\cite{sunderhauf2023density} use 5 similar views for initialization and plans NBVs up to 30 views. In contrast to these fixed constraints, we adaptively predict the number of required views based on the complexity of the object to be reconstructed, enabling us to effectively allocate the measurement budget.


\section{System Overview}
\label{sec_system}

\begin{figure}[!t]
\centering
\includegraphics[width=1.0\columnwidth]{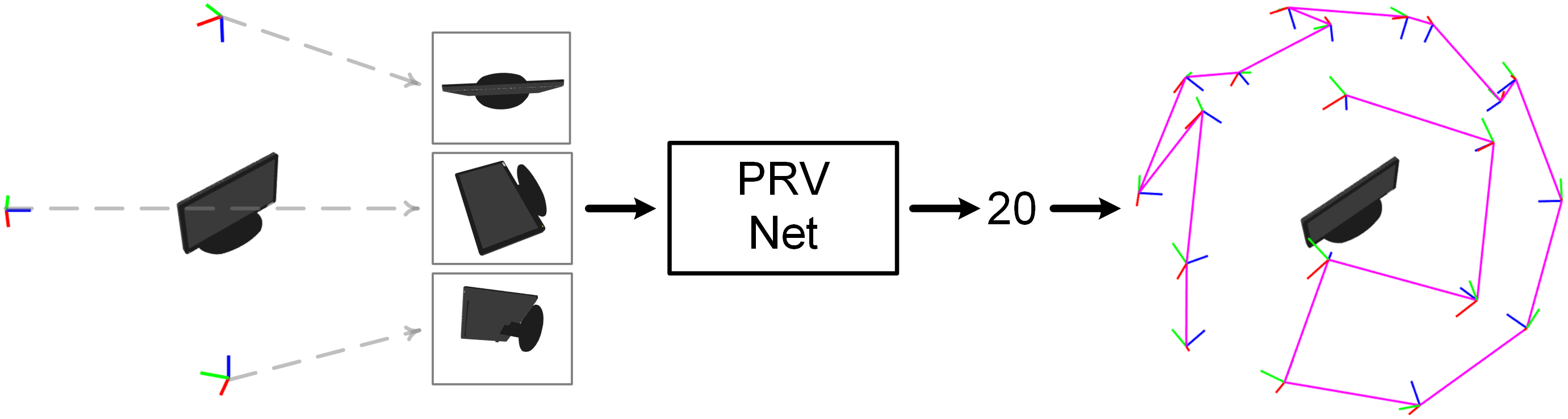}
\caption{
An example of our online workflow given three initial views. The selected initial views (top, left, and front) are represented by red-green-blue axes. The robot takes these images and stops at the top view. We input these images into our PRVNet to obtain a predicted number of views for the reconstruction~(20 in this example). Based on this, we generate the Tammes view space \cite{lai2023iterated} of size 20 and the purple global path for the robot to execute. 
} 
\label{fig_workflow}
\vspace{-0.2cm}
\end{figure}

Our goal is to actively reconstruct the NeRF representation of an unknown object in a tabletop scenario. This reconstruction is accomplished by utilizing a series of posed RGB images captured from various sensor views guided by a robotic arm. Fig.~\ref{fig_workflow} shows the workflow of the online phase of our object reconstruction system.

The online phase begins with the robotic acquisition of three images from initial views: top, left or right, and front or back, which encompass crucial information, such as size and texture, about the object on the tabletop. The setup of initial views is confirmed in the ablation study presented in Sec.~\ref{sec_exp_network}. The robot traverses three views and stops at the top view. Subsequently, we input these images into our PRVNet to predict the object-specific required number of views for reconstructing the object as detailed in Sec.~\ref{sec_predict}. This prediction determines the generation of a Tammes view space surrounding the object, and a global path is calculated for traversing these views as explained in Sec.~\ref{sec_system_vpp}. The robot then navigates to each view according to the global path, capturing images and saving them along with their corresponding view poses. In the offline phase, these posed images, including the three initial measurements, are used for the NeRF reconstruction.

\section{View Space and Path Planning} 
\label{sec_system_vpp}

We assume a hemispherical view space on the tabletop with views pointing to the center of the hemisphere, as often considered in active object reconstruction approaches~\cite{sunderhauf2023density,lee2022uncertainty,jin2023neu, pan2022activenerf, pan2022scvp, ran2023neurar}. The position of each view is defined by Tammes problem~\cite{lai2023iterated} that solves the task of placing a given number of points on a sphere to maximize the minimum distance between them. 

Our global path planning method solves the problem of connecting all views in the Tammes view space. We generate the optimal global path by solving the shortest Hamiltonian path problem on a graph, which is similar to the traveling salesman problem (TSP) but without returning to the starting node. As Gurobi Optimizer~\cite{gurobi2021gurobi} efficiently resolves TSP (less than 100 nodes) within seconds, we introduce a virtual starting node to convert the Hamiltonian path problem into a TSP scenario and efficiently obtain the final robot global path. An illustration of the global path is given in Fig.~\ref{fig_workflow}. The view-to-view local path follows the concept of avoiding the object as an obstacle on the tabletop as fully defined in~\cite{pan2023one}.


\section{Predicting Required Number of Views in NeRF}
\label{sec_predict}

This section presents our novel PRVNet, which is designed to adaptively determine the required number of views for a specific object. For network training, we generate a dataset consisting of individual objects and their required number of views. Given initial measurements of an object as input, PRVNet is trained under the supervision of the corresponding view-number label. 

\subsection{Object-Specific Required Number of Views} \label{sec_predict_complexity}

\begin{figure}[!t]
\centering
\includegraphics[width=1.0\columnwidth]{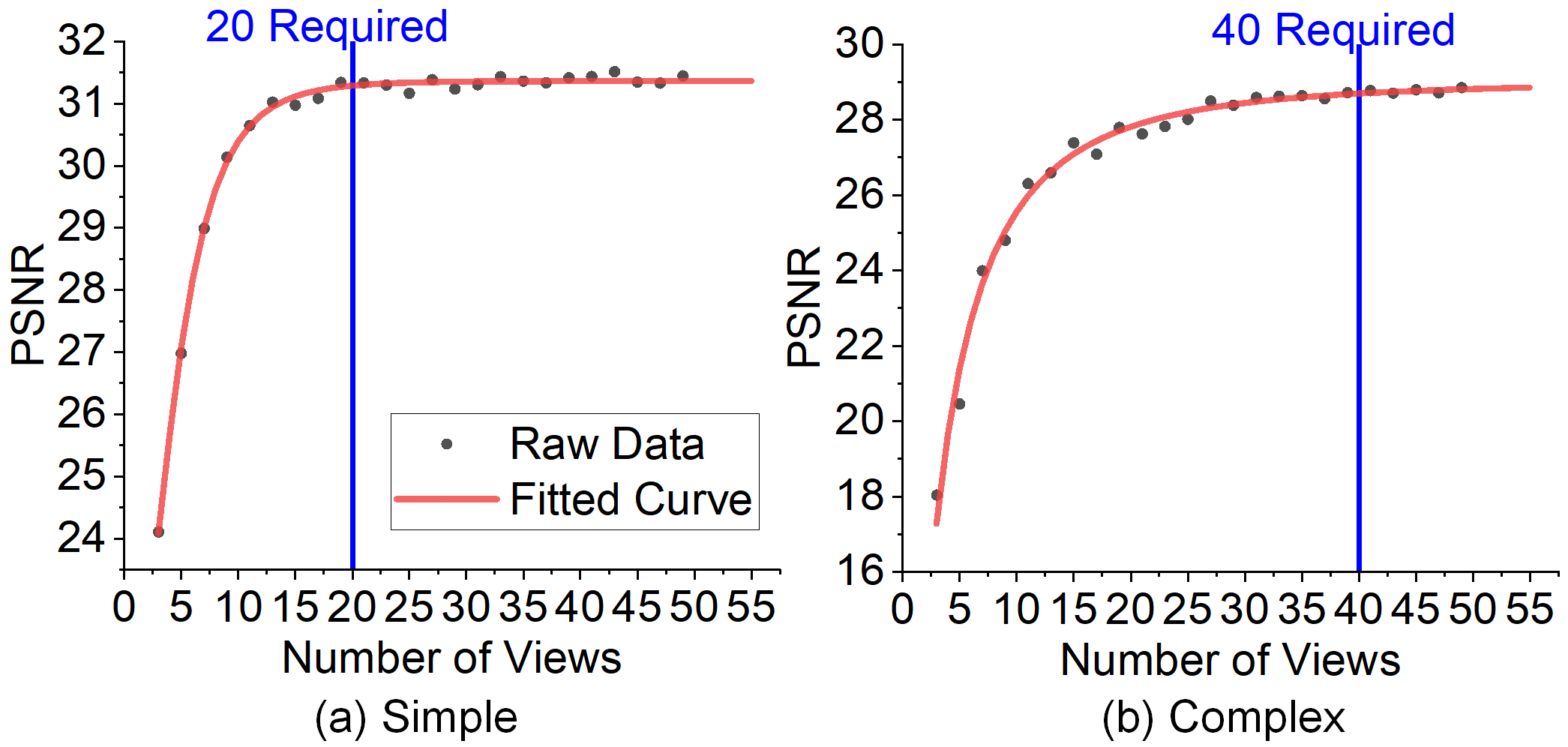}
\caption{
An example of quantitative analysis of the required number of views on different object complexity: (a) a simple object, (b) a complex object. Each black point is a pair of $(v, \mbox{PSNR})$, which means a NeRF trained under a view space of size $v$, and images from 100 test Tammes views are rendered to report an average PSNR value. The red curve $C_o$ is fitted to these data points to determine the $v^\ast$ based on its gradient. The blue lines suggest that for a simple object, we achieve a satisfactory result with only 20 views, whereas a complex object necessitates 40 views.
} 
\label{fig_object_complexity}
\vspace{-0.2cm}
\end{figure}

Our proposed approach is based on the key insight that as the object complexity increases, a larger number of views is necessary to obtain a good NeRF representation. To quantitatively study this relationship, we plot the PSNR value for a specific object over the number of views $v \in \mathbb{N}^{+} $, ranging from 3 to 50 at intervals of 2. Fig.~\ref{fig_object_complexity} illustrates the plots for two example objects. While the PSNR values may fluctuate due to the inherent training randomness in CUDA~\cite{muller2022instant}, we observe that with a higher count of views, the rate of PSNR growth diminishes to zero for a specific object. We exploit these convergence trends to assess the complexity levels of different objects.

\begin{figure}[!t]
\centering
\includegraphics[width=1.0\columnwidth]{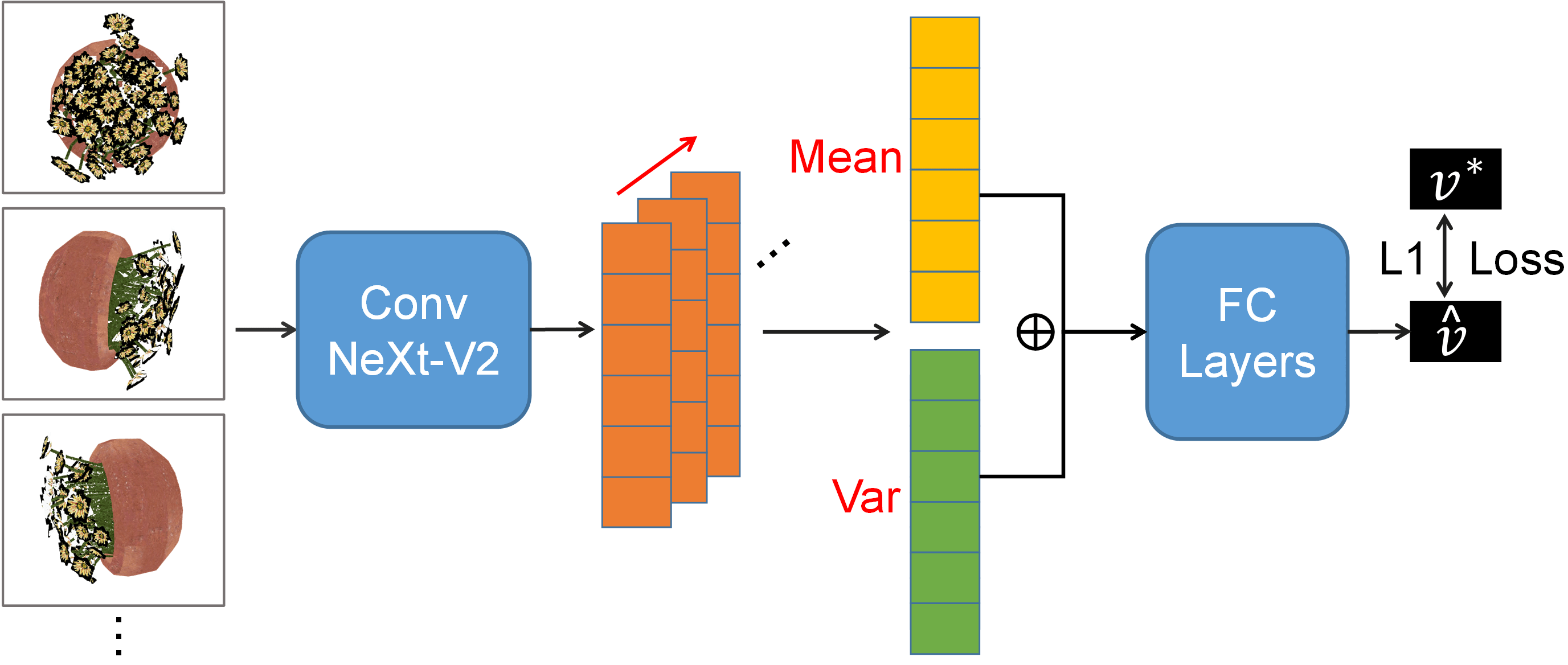}
\caption{
PRVNet architecture: We use the state-of-the-art ConvNeXt-V2~\cite{woo2023convnext} as the backbone to extract features from each image. The red arrow indicates the calculation of mean and variance across the batch dimension. $\bigoplus$ represents the concatenation operation. L1 loss is employed for network training.
} 
\label{fig_prv_network}
\vspace{-0.5cm}
\end{figure}

\subsection{Labeling of Required Number of Views} \label{sec_predict_curve}

To ensure the monotonic increase of the growth rate used for labeling, we use a curve-fitting approach to mitigate the impact of fluctuations. Given that the growth rate typically follows a skewed distribution (decreasing as the number of views increases), a log-normal distribution is often assumed~\cite{crow1987lognormal}. Consequently, the raw data can be fitted to a cumulative distribution function, denoted as $\mathit{C_o}$, as depicted in Fig.~\ref{fig_object_complexity}. When the gradient of the growth rate falls below a certain small threshold $\alpha$, we deduce that a required number of views $v^\ast \in \mathbb{N}^{+}$ is sufficient for an object $o$ to achieve a satisfactory NeRF representation:
\begin{equation}
\label{equ_vb_ast}
\begin{aligned}
\mathit{v^\ast = \argmin(v),\ \mathrm{s.t.}\ C_o(v+1)-C_o(v)<\alpha} \, .
\end{aligned}
\end{equation}
Once the required number of views label $v^\ast$ is computed for an object $o$, we generate the supervision pair $(I_o, v^\ast)$ for the network training, where $I_o$ is the list of images taken from several initial views around object $o$.
\subsection{Learn to Predict the Required Number of Views} \label{sec_predict_prv}

Our PRVNet is a regression network that takes several images $I_o$ as input. To enable the PRVNet to process multiple image inputs and learn multi-view information, we devised a network architecture illustrated in Fig.~\ref{fig_prv_network}. The output of the last Sigmoid activation is converted to the range of required numbers of views in our dataset by linear mapping and constant offset to acquire final prediction $\hat{v} \in \mathbb{R}^{+}$. We use L1 loss to enforce our PRVNet to predict a value close to the ground truth label. During deployment, $\hat{v}$ is rounded to an integer for configuring Tammes view space.

\section{Experiments}
\label{sec_exp}

Our experiments are designed to support the claim that our method can achieve fast online data collection for high-quality NeRF reconstruction of an unknown object by predicting the required number of views.

\begin{figure}[!t]
\centering
\includegraphics[width=1.0\columnwidth]{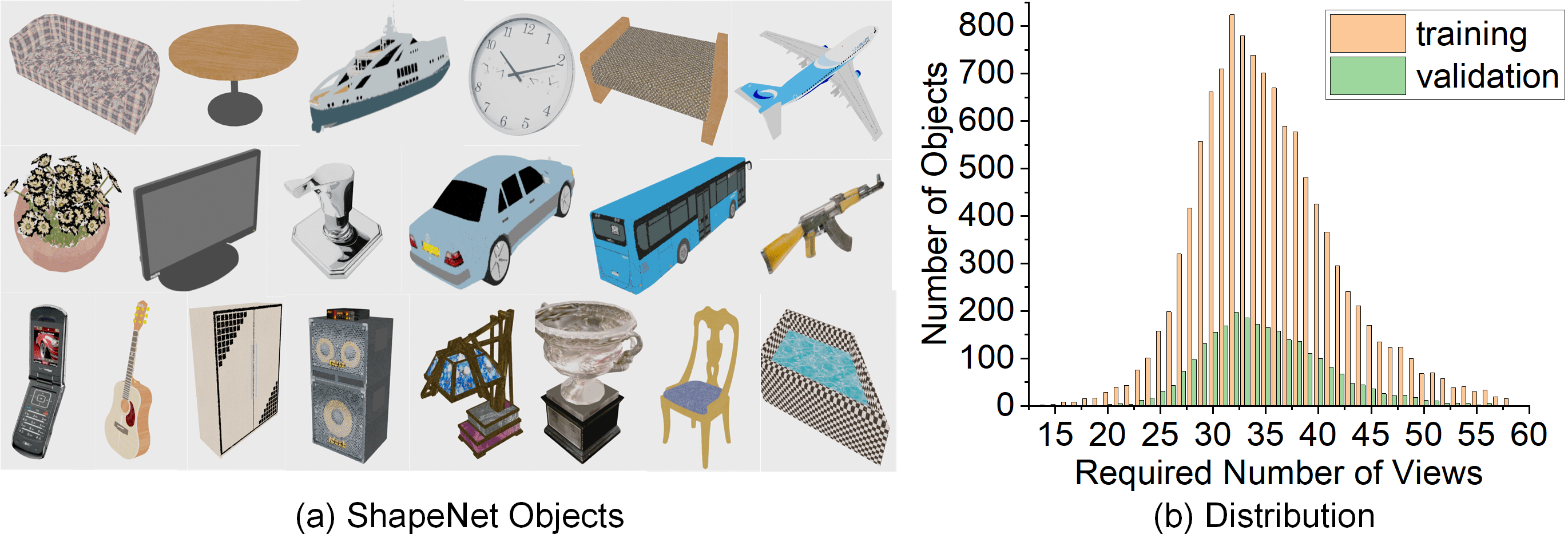}
\caption{
3D model and required number of views datasets: (a) textured examples of the top 20 classes in ShapeNet~\cite{shapenet2015}; (b) training and validation set distributions over the required number of views.
} 
\label{fig_dataset_distribution}
\vspace{-0.2cm}
\end{figure}

\begin{figure}[!t]
\centering
\includegraphics[width=1.0\columnwidth]{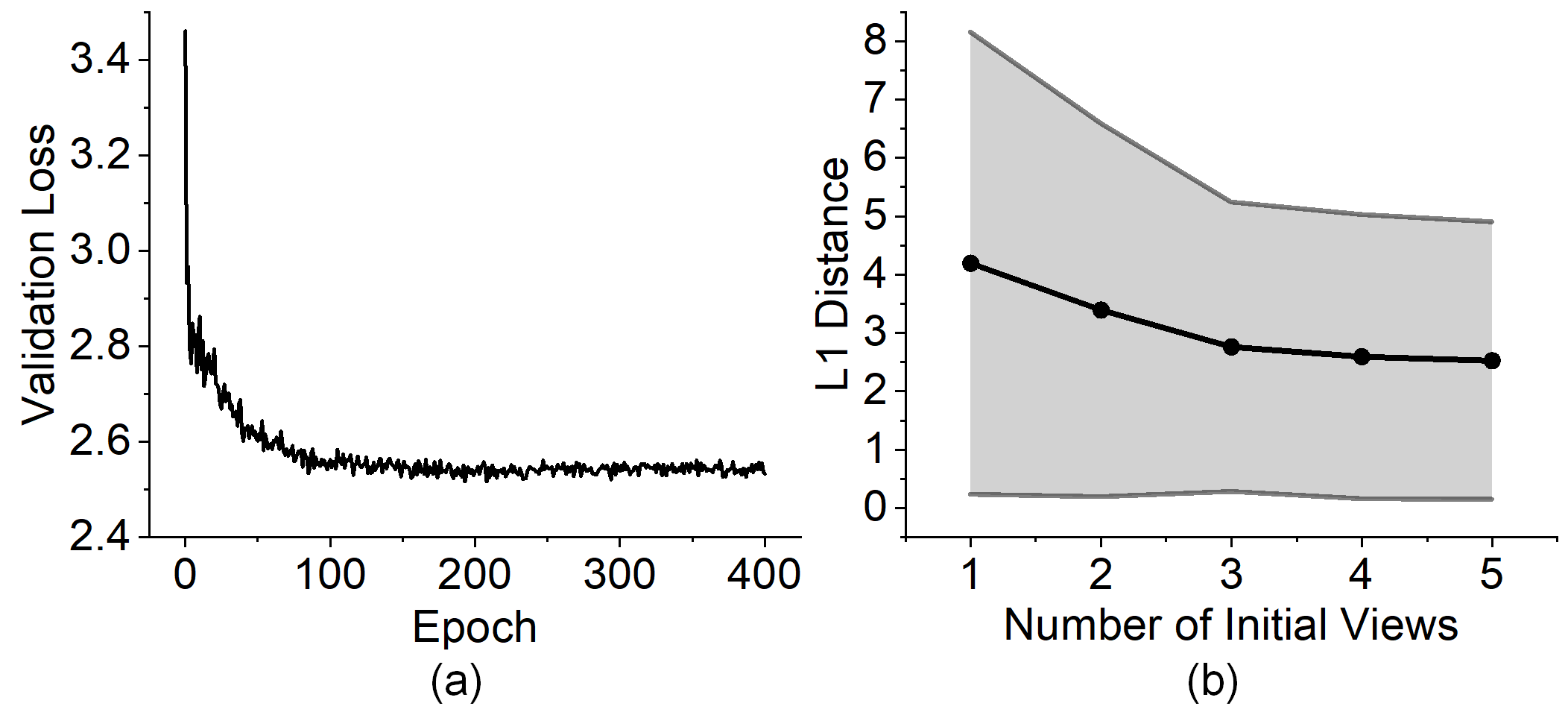}
\caption{
PRVNet training: (a) validation loss over epochs, (b) L1 distance with its standard deviation over the number of initial views (the black curve with the gray error bar).
} 
\label{fig_network_training}
\vspace{-0.5cm}
\end{figure}

\begin{table*}[!t]
\centering
\resizebox{0.75\textwidth}{!}{%
\begin{tabular}{|c|c|c|c|c|c|c|c|}
\hline
Method   & Required Views & PSNR Difference\,$\downarrow$ & SSIM Difference\,$\downarrow$ & Movement Difference\,(m)\,$\downarrow$ \\ \hline
GT Label & 34.94\,±\,5.70 & 0                       & 0                 & 0               \\ \hline
Mode     & 32             & 0.1641\,±\,0.1551           & 0.002163\,±\,0.002370 & 0.3070\,±\,0.3117   \\ \hline
Median   & 34             & 0.1511\,±\,0.1630           & 0.002248\,±\,0.002397 & 0.2896\,±\,0.2624   \\ \hline
Mean     & 35             & 0.1624\,±\,0.1581           & 0.002086\,±\,0.002278 & 0.2913\,±\,0.2458   \\ \hline
PRVNet\,(Proposed)   & 35.77\,±\,5.24    & \textbf{0.1390}\,±\,0.1588           & \textbf{0.001817}\,±\,0.001928 & $^\star$\textbf{0.1988}\,±\,0.2032   \\ \hline
\end{tabular}
}
\caption{Comparison to statistic methods. We report five different methods to compute the required number of views: Ground Truth Label (GT Label), Mode 32, Median 34, Mean 35 shown in Fig.~\ref{fig_dataset_distribution}(b), as well as our PRVNet. PSNR and SSIM are computed from 100 novel views. The Movement/PSNR/SSIM Difference stands for the absolute difference from the GT Label. Each value reports the average mean and standard deviation on 250 random objects from the validation set. The star indicates significant results against statistic methods according to the paired \textit{t}-test with a \textit{p}-value of 0.05. As can be seen, our PRVNet output aligns most closely with the GT Label, indicating a good prediction of the required number of views based on object complexity.
}
\label{tab_statistic}
\vspace{-0.3cm}
\end{table*}

\begin{figure*}[!t]
\centering
\includegraphics[width=1.0\textwidth]{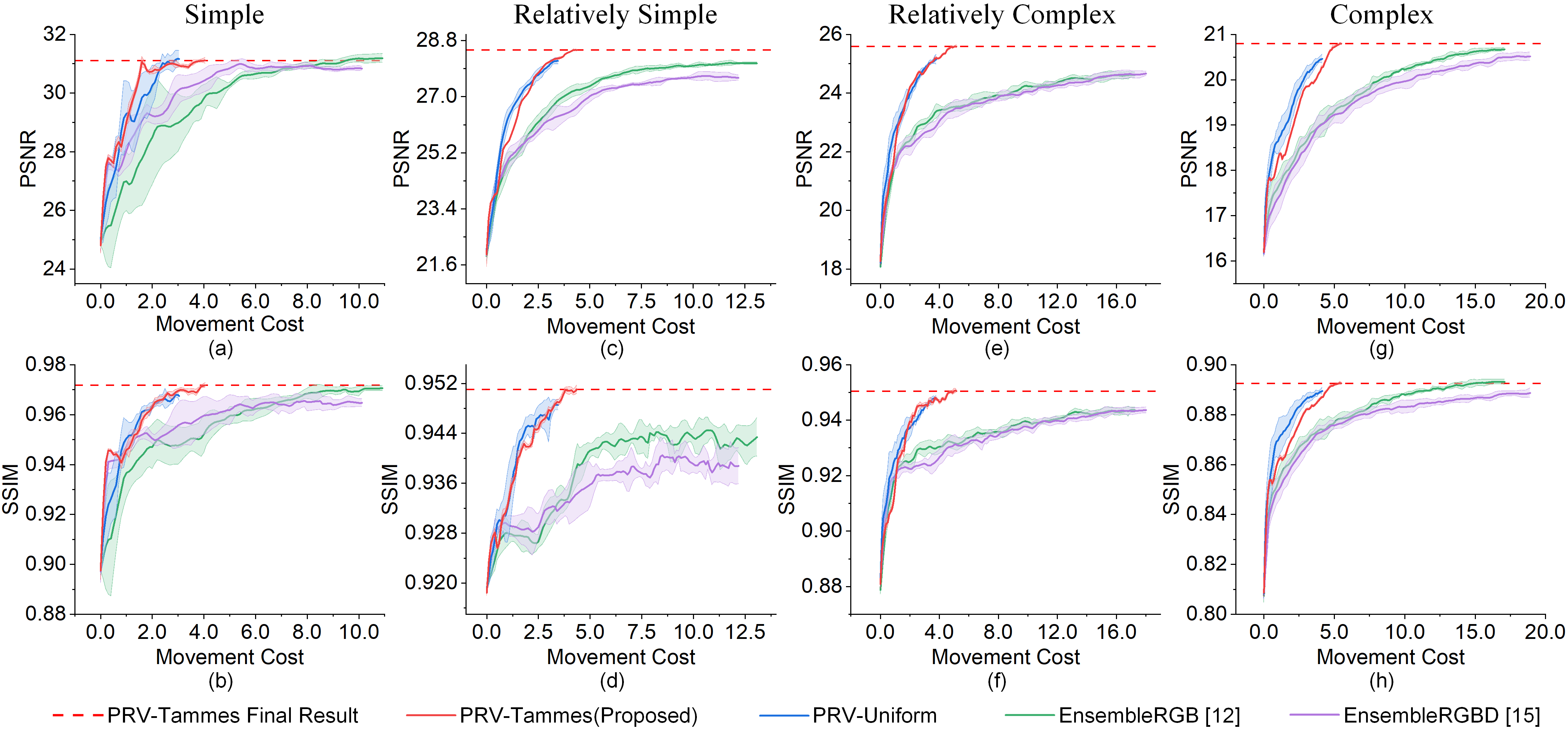}
\caption{
Comparison of view planning results. We report four objects with different complexity, \ie required number of views in ground truth labels varying from small to large: (a),\,(b) Simple 20; (c),\,(d) Relatively Simple 30; (e),\,(f) Relatively Complex 40; (g),\,(h) Complex 50. PSNR and SSIM are computed from 100 novel views. Each row shows PSNR and SSIM, respectively, over the movement cost, which is computed by linear interpolation with the interval 0.1 because only iterative data are available. Each planner is tested 5 times and its average mean with standard deviation (the error bar) is reported. As can be seen, (1) our PRV-based methods achieve superior or comparable PSNR/SSIM while requiring less movement cost than other baselines, particularly for more complex objects; (2) our Tammes view configuration achieves better reconstruction with smaller standard deviations than uniformly sampled configurations.
} 
\label{fig_showcase}
\vspace{-0.65cm}
\end{figure*}

\subsection{Dataset Generation} \label{sec_exp_dataset}

\textbf{Object 3D Model Dataset.} The capabilities of our method depend on a well-trained PRVNet. We generate the required number of views dataset on the ShapeNet 3D mesh model dataset~\cite{shapenet2015} that contains different classes of objects. Given the imbalanced distribution of objects across various classes within the ShapeNet, we consider a maximum of 1,200 objects per class for the top 20 classes as shown in Fig.~\ref{fig_dataset_distribution}(a). On the other hand, texture information is important to object complexity. We therefore only consider 3D models with textures and use a sampling method~\cite{lazzarotto2022sampling} to ensure the visual result is the same as the original mesh.

\textbf{Required Number of Views Dataset.} We perform virtual imaging of resolution $1280\times720$ px on object 3D models from different views in a simulation environment. Considering a real-world tabletop environment, we set the radius of view spaces to 0.3\,m. Since object size can also be interpreted as part of the object complexity, we randomize the object size from 0.07\,m to 0.12\,m as data augmentation. After obtaining these ground truth images from 3-50 view spaces, we train a NeRF for each object and required number of view pairs to get the PSNR data plots for fitting $C_o$ as discussed in Sec.~\ref{sec_predict_complexity}. The training and rendering of these NeRFs are implemented using Instant-NGP \cite{muller2022instant} with a training step of 2,500 on a cluster of 8 NVIDIA A100 Tensor Core GPUs. The Orthogonal Distance Regression method in OriginPro~\cite{originlab2021originpro} is used for curve fitting. In total, we label 13,789 objects with the required number of views under $\alpha=0.02$. We employ an 8/2 ratio to randomly partition our dataset into training and validation set as shown in Fig.~\ref{fig_dataset_distribution}(b). 

\begin{table*}[!t]
\centering
\resizebox{0.80\textwidth}{!}{%
\begin{tabular}{|c|c|c|c|c|} 
\hline
Method & PSNR\,$\uparrow$ & SSIM\,$\uparrow$ & Movement Cost\,(m)\,$\downarrow$ & Planning Time\,(s)\,$\downarrow$ \\
\hline
EnsembleRGB~\cite{lin2022active} & 26.96\,±\,2.86 & 0.9419\,±\,0.0879 & 12.719\,±\,2.510 & 2536.5\,±\,500.1 \\
\hline
EnsembleRGBD~\cite{sunderhauf2023density} & 27.09\,±\,2.17 & 0.9526\,±\,0.0239 & 12.458\,±\,2.440 & 2600.3\,±\,521.6 \\			
\hline
PRV-Uniform & 27.49\,±\,2.25 & 0.9562\,±\,0.0230 & \textbf{3.336}\,±\,0.316 & 0.687\,±\,0.039 \\
\hline
PRV-Tammes\,(Proposed) & \textbf{27.84}\,±\,2.25 & \textbf{0.9577}\,±\,0.0229 & 4.589\,±\,0.372 & \textbf{0.605}\,±\,0.003 \\
\hline
\end{tabular}
}
\caption{Comparison of final reconstruction results. We report the metrics after all images are collected. PSNR and SSIM are computed from 100 novel views. The movement cost and planning time are total sum values during the online reconstruction. Each value is reported as the averaged mean and standard deviation on 50 random objects from the validation set. Two ensemble baselines~\cite{lin2022active,sunderhauf2023density} are assumed to be fully paralleled and the reported time is divided by the number of ensembles. Note that the standard deviations primarily arise from the complexity of different objects. As can be seen, the proposed PRV-Tammes method is highly efficient in terms of movement cost and planning time along with better quality. Note that PRV-Uniform requires slightly higher planning time due to online global path computation, whereas the proposed PRV-Tammes benefits from the use of a look-up table of pre-calculated Tammes view spaces.
}
\label{tab_vp}
\vspace{-0.5cm}
\end{table*}

\subsection{Network Training} \label{sec_exp_network}

\textbf{Implementation Details and Parameters.} We use the tiny model for ConvNeXt-V2 and set the output feature layer as 1,000. The fully connected (FC) layers are then sized as [1,000,\,500,\,250,\,100,\,1]. The output of PRVNet is remapped to [13,\,58] as the range of views in our dataset shown in Fig.~\ref{fig_dataset_distribution}(b). The batch size is set to 64, the base learning rate is set to 0.00015, and the weight decay is set to 0.05. The pre-trained weights of ConvNeXt-V2 on ImageNet-1K are used in our PRVNet for better feature extraction. The size of $I_o$ is set to the number of initial views (top, left, right, front, and back) used for training. We train our PRVNet for 400 epochs on 8 A100 GPUs. The validation loss over epochs is shown in Fig.~\ref{fig_network_training}(a). We save the network with the smallest L1 loss on the validation set as the final result. 

\textbf{Ablation Study on Initial Views.} Fig.~\ref{fig_network_training}(b) reports the results from different numbers of initial views on the validation set. Although the best results could be achieved with five views, the setup of three views is stable enough to use. We finally chose $|I_o|=3$ as input for the network to improve the reconstruction speed.

\textbf{Comparison to Statistic Methods.} We perform a comparison with the basic statistics, \ie the number of views for each object, as shown in Table~\ref{tab_statistic}. From the results, we confirm that the PSNR, SSIM (Structural Similarity Index)~\cite{mildenhall2021nerf}, and movement cost differences (with respect to the ground truth results) of the PRVNet are smaller than the statistical methods. This means that the PRVNet effectively predicts the appropriate required number of views for objects of varying complexity, \ie giving an object-specific prediction.

\subsection{Evaluation of View Planning} \label{sec_exp_vp}

\textbf{Baselines and Metrics.} We compare our PRV-based view planning method (PRV-Tammes) with two uncertainty-based NBV methods (EnsembleRGB~\cite{lin2022active} and EnsembleRGBD~\cite{sunderhauf2023density}). We set a planning view space of size 540 for the baselines. The resolution of ensembles is set to $90\times45$ to have a rapid uncertainty rendering. In addition, we also perform an ablation study on our Tammes configuration by replacing it with uniformly sampled views (PRV-Uniform), \ie random sampling views from the planning view space with the number of the PRVNet output. The global path planning is also used for PRV-Uniform. We evaluate the methods on an i7-12700H CPU and an Nvidia RTX3060 Laptop GPU to represent deployment scenarios. We use PSNR and SSIM to evaluate the quality of NeRF representations. The movement cost (accumulated Euclidean distance) and planning time (the sum of inference time and path planning time) are used to evaluate the efficiency.

\textbf{Setup and Results.} For a fair comparison, the same three initial views (top, left, and front) are configured for each method, and the number of views is set as the same as the output from the PRVNet. Note that, unlike the previous comparison of statistics, the images of the initial views are also included in the NeRF training. Two sources of randomness influence the planning results: (i)~the Instant-NGP training process; (ii)~the planning methods. We thus perform multiple trials for each planner on four objects from simple to complex to explore these randomnesses in Fig.~\ref{fig_showcase}, as well as the final reconstruction results on more object cases in Table~\ref{tab_vp}. From the results, we confirm that: (1)~The proposed PRV-Tammes method achieves higher or similar PSNR/SSIM quality within less movement cost than other baselines, especially for the more complex objects. (2)~Our PRV-based methods require very little planning time, \ie inferring with the PRVNet once, compared to iterative baselines, where retraining NeRF is required between planning steps. (3)~PRV-Uniform method also achieves high reconstruction efficiency but lower final PSNR/SSIM and is less stable (larger standard deviation in Fig.~\ref{fig_showcase}) than our Tammes configuration.

\begin{figure}[!t]
\centering
\includegraphics[width=0.95\columnwidth]{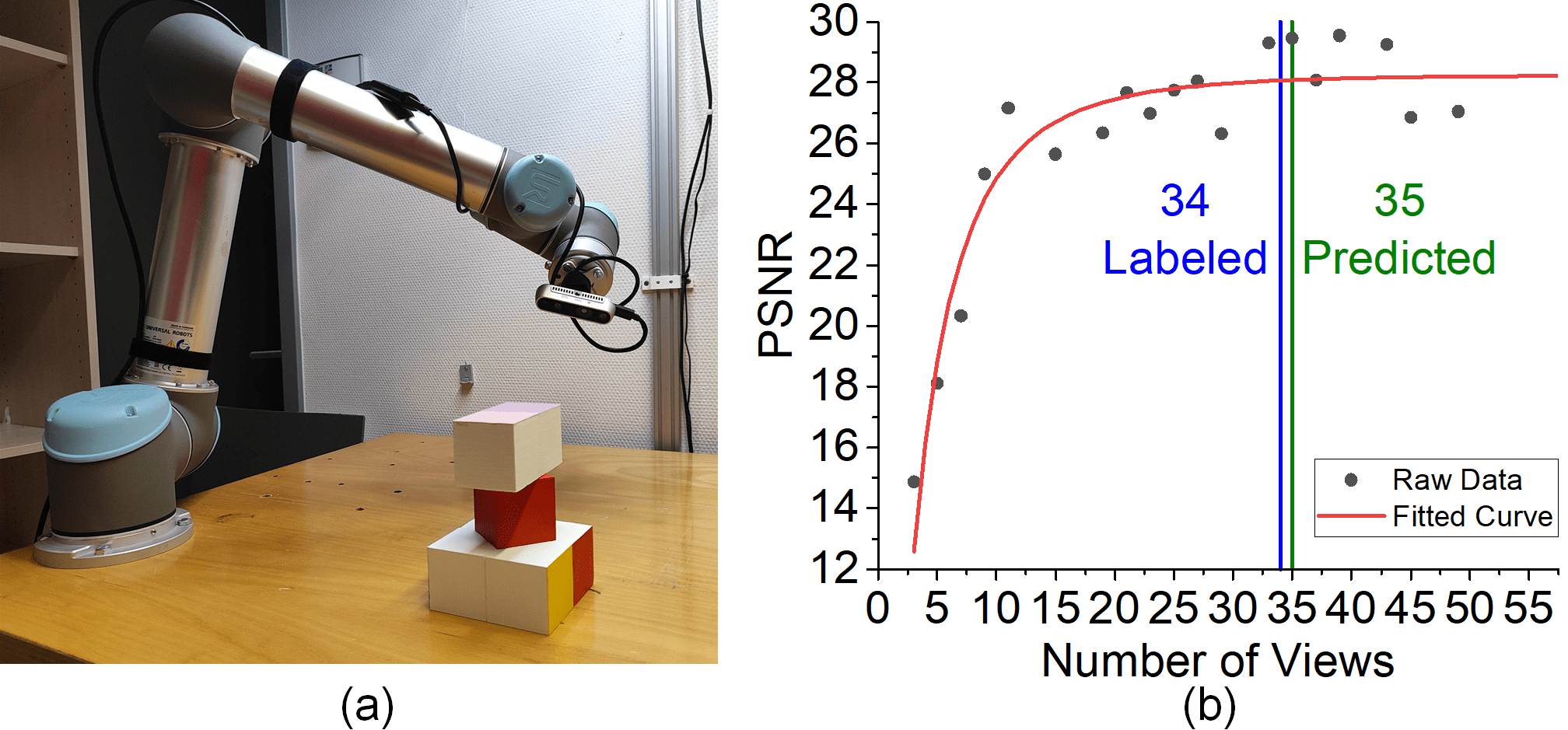}
\caption{
Real-world reconstruction: (a) robot environment and the object, (b) generalization test. The required number of views labeled is computed by the fitted PSNR curve as 34 (blue line). Our PRVNet takes three initial images and outputs 35 (green line). The small difference in the required view number indicates that our PRVNet has a reasonable sim-to-real performance.
} 
\label{fig_realworld}
\vspace{-0.5cm}
\end{figure}

\subsection{Real-World Reconstruction} \label{sec_exp_rw}

\textbf{Setup.} We validate our approach in a real-world environment using a UR5 robot arm with an Intel Realsense D435 camera mounted on its end-effector (only the RGB optical camera is activated). ROS~\cite{koubaa2017robot} and MoveIt~\cite{chitta2016moveit} are used for robotic motion planning.

\textbf{Generalization and Reconstruction Test.} We collect real-world images from different Tammes view spaces of an object to compute the label of the required number of views introduced in Sec.~\ref{sec_predict_curve}. The experimental environment and data are shown in Fig.~\ref{fig_realworld}. The online data collection process and final reconstruction results are presented in the accompanying video at \url{https://youtu.be/LoQGOR3S1Fw}. From the results, we confirm that: (1)~Our PRVNet can generalize to real-world environments, and (2)~our PRV-based view planning achieves fast online image collection and good NeRF reconstruction quality.

\section{Conclusion and Discussion}
\label{sec_con}

In this paper, we present a novel non-iterative pipeline for active NeRF reconstruction using the prediction of the required number of views and the Tammes configuration for view pose generation. We propose PRVNet trained on our new dataset consisting of objects of different complexities to predict the required number of views. We leverage the network output to plan a globally connected path representing the minimum travel distance. Our experiments show that our view planning using the PRVNet prediction achieves a higher efficiency in terms of movement cost and competitive quality in reconstruction compared to state-of-the-art baselines. Our pipeline holds promise for robotic applications, particularly in tasks like volume estimation, grasping, and real-time object reconstruction during online missions. Instead of using a fixed Tammes view configuration, our future work would consider adaptive view configurations according to specific objects to be reconstructed.


\bibliographystyle{IEEEtranN}
\footnotesize
\bibliography{bibliography}

\end{document}